%% file: main.tex
\documentclass[english]{article}
\usepackage{graphicx} 
\input{preambles}

\title{Towards Anthropomorphic Conversational AI \\
Part I: A Practical Framework}
\date{}

\author{
Fei Wei, Yaliang Li, Bolin Ding
\\
\\
\small{Alibaba Group}
}
\begin{document}

\maketitle

\begin{abstract}
Large language models (LLMs), due to their advanced natural language capabilities, have seen significant success in applications where the user interface is usually a conversational artificial intelligence (AI) agent and engages the user through multi-round conversations. However, many scenarios require the agents to exhibit stronger social and conversational intelligence and demonstrate more human-like (anthropomorphic) reactions. This is an aspect that foundational LLMs have yet to fully address such that a single call of foundational models might be insufficient.

To bridge this gap, we propose a two-stage solution. In this work, we focus on the first stage, introducing a multi-module framework designed to replicate the key aspects of human intelligence involved in conversations. This framework comprises thinking modules for reasoning, resource modules for managing knowledge and external information, and response modules for generating contextually appropriate interactions. With all the modules cooperating, the framework would empower the agents to provide a better human-like conversation experience.
In the second stage of our approach, these conversational data—after filtering and labeling—can serve as training and testing data for reinforcement learning, enabling AI to better capture human preferences. This stage is left for future work.

In our experiments, volunteers engaged in over 3,000 rounds of conversation with the same AI character powered by a standalone LLM and our framework which integrates the same LLM. A separate group of evaluators rated the conversation samples, revealing that our framework significantly enhanced the AI’s social and conversational intelligence, even without fine-tuning the LLM. Additionally, our framework generated responses enriched with comprehensive reasoning.\footnote{The source code is under review and will be available soon.}

\end{abstract}

\input{sections/introduction}

\input{sections/preliminaries}

\input{sections/system}

\input{sections/experiments}

\input{sections/conclusion}

\clearpage
\bibliographystyle{plain}
\bibliography{ref}

\end{document}

%% file: preambles.tex
\usepackage[most]{tcolorbox}
\usepackage{graphicx} 
\usepackage{caption}
\usepackage{subcaption}
\usepackage{xcolor}
\usepackage{xspace}
\usepackage{booktabs}
\usepackage{url}
\usepackage{multirow}
\usepackage[linesnumbered, boxed]{algorithm2e}
\usepackage{amsmath, amsthm}
\usepackage{listings}

\usepackage[utf8]{inputenc} 
\usepackage[T1]{fontenc}    
\usepackage{hyperref}       
\usepackage{url}            
\usepackage{amsfonts}       
\usepackage{nicefrac}       
\usepackage{microtype}      

\usepackage{amssymb}

\usepackage{geometry}
\usepackage[numbers]{natbib} 
\usepackage{babel}
\usepackage{siunitx}
\usepackage{bm}
\usepackage{bbm}
\usepackage{appendix}

\definecolor{codegreen}{rgb}{0,0.6,0}
\definecolor{codegray}{rgb}{0.5,0.5,0.5}
\definecolor{codepurple}{rgb}{0.58,0,0.82}
\definecolor{backcolour}{rgb}{0.95,0.95,0.92}

\definecolor{eclipseBlue}{RGB}{42,0.0,255}
\definecolor{eclipseGreen}{RGB}{63,180,95}
\definecolor{eclipsePurple}{RGB}{175,0,25}
\definecolor{codewhite}{rgb}{0.70,0.70,0.70}

\definecolor{moderator}{HTML}{e3716e}
\definecolor{player1}{HTML}{bdb5e1}
\definecolor{player2}{HTML}{7ac7e2}
\definecolor{player3}{HTML}{f7df87}
\definecolor{player4}{HTML}{54beaa}
\definecolor{player5}{HTML}{2983b1}
\definecolor{player6}{HTML}{eca680}

\lstdefinestyle{mystyle}{
    backgroundcolor=\color{backcolour},   
    commentstyle=\color{codegreen},
    keywordstyle=\color{magenta},
    numberstyle=\tiny\color{codegray},
    stringstyle=\color{codepurple},
    basicstyle=\ttfamily\footnotesize,
    breakatwhitespace=false,         
    breaklines=true,
    breakindent=-3.5pt,
    captionpos=b,                    
    keepspaces=true,                 
    numbers=left,                    
    numbersep=5pt,                  
    showspaces=false,                
    showstringspaces=false,
    showtabs=false,                  
    tabsize=4,
    postbreak={
    	\mbox{
    		\lst@linebreakbgrd
    		\rotatebox[y=0.7ex]{180}{\color{black}$\Lsh\,$}
    	}
    },
}

\lstdefinelanguage{Dialog}{
	backgroundcolor=\color{backcolour},   
	keywordstyle=\color{magenta},
	numberstyle=\tiny\color{codegray},
	basicstyle=\ttfamily\footnotesize,
	breakatwhitespace=false,         
	breaklines=true,   
    breakindent=-3.5pt,
	captionpos=b,                    
	keepspaces=true,                 
	numbers=left,                    
	numbersep=5pt,                  
	showspaces=false,                
	showstringspaces=false,
	showtabs=false,                  
	tabsize=4,
	morecomment = [l][\color{eclipseGreen}\bfseries]{Assistant:},
    morecomment = [s][\color{eclipseGreen}\bfseries]{1.}{more.},
    morecomment = [l][\color{eclipseBlue}\bfseries]{User:},
}

\lstdefinelanguage{Werewolf}{
	backgroundcolor=\color{backcolour},   
	keywordstyle=\color{magenta},
	numberstyle=\tiny\color{codegray},
	basicstyle=\ttfamily\footnotesize,
	breakatwhitespace=false,         
	breaklines=true,   
    breakindent=-3.5pt,
	captionpos=b,                    
	keepspaces=true,                 
	numbers=left,                    
	numbersep=5pt,                  
	showspaces=false,                
	showstringspaces=false,
	showtabs=false,                  
	tabsize=4,
    morecomment = [l][\color{player1}\bfseries]{Player1:},
     morecomment = [l][\color{player2}\bfseries]{Player2 :},
     morecomment = [l][\color{player3}\bfseries]{Player3 :},
     morecomment = [l][\color{player4}\bfseries]{Player4 :},
     morecomment = [l][\color{player5}\bfseries]{Player5 :},
     morecomment = [l][\color{player6}\bfseries]{Player6 :},
     morecomment = [s][\color{moderator}\bfseries]{Moderator :}{\}},
}

%% file: sections/introduction.tex
\section{Introduction}

In recent years, large language models (LLMs), such as GPT, Gemini, Llama, Deepseek, and Qwen \cite{achiam2023gpt, team2023gemini, dubey2024llama, liu2024deepseek, yang2024qwen2}, have demonstrated remarkable capability in conversational artificial intelligence (AI) systems, which are designed to process, understand, and generate human-like conversations through text or speech, and enable meaningful interactions with users in various applications such as chatbots, virtual assistants, and customer service automation.

In those applications, the user interface (UI) is usually represented as a virtual character (agent), which is sometimes granted with a persona, and interacts with the users through multi-turn conversations.
The ability of LLMs to generate fluent, contextually relevant responses makes them well-suited for conversational AI, enhancing user experience and engagement.

Anthropomorphism, a classical concept in psychology, refers to the attribution of human traits, emotions, or intentions to non-human entities and has attracted increasing attention in the literature of AI research \cite{salles2020anthropomorphism}.
In the context of conversational AI, anthropomorphism primarily focuses on granting AI systems the capability to think in a human-like manner and communicate using natural, human-like language \cite{salles2020anthropomorphism}.
Studies suggest that anthropomorphic AI can enhance communication efficiency and quality, foster positive social relationships \cite{hohenstein2023artificial}, improve user satisfaction and engagement \cite{jo2024effects}, and introduce new social dynamics, where users may perceive AI as partners, advisors, or even emotional companions \cite{guzman2020artificial}. 
Given these benefits, human-like (anthropomorphic) behavior of conversational AI is particularly desirable in highly interactive and social applications, such as education \cite{ji2023systematic}, customer service \& support \cite{adam2021ai}, and healthcare \cite{d2020ai}, where the agent is expected to show more human-likeness rather than only respond with fluent and correct natural language.

However, achieving anthropomorphic AI in conversational systems presents two major challenges: a relatively easier one - expressing like a human - ensuring AI-generated text mimics natural spoken language, and a more challenging one - thinking like a human - enabling AI to exhibit reasoning and contextual awareness similar to human cognition.

\subsection{Expressing like a Human}

A significant challenge in LLM-based conversation is ensuring that AI-generated text feels natural and conversational, rather than overly structured or formal. 
Since textual data—including books, articles, and research papers—serves as the primary training resource, LLMs are inherently better at producing well-structured written text than informal spoken dialogue \cite{brown2020language, radford2018improving}.
As a result, most pre-trained LLMs struggle with: i) generating natural spoken language, which tends to be informal, fragmented, and context-dependent, ii) matching conversational tone and style, as AI-generated responses often sound overly formal or mechanical, iii) adjusting for conversational fluidity, where free-flowing dialogue and casual speech patterns are harder for LLMs to replicate.
For example, users often report that LLMs produce responses that feel too formal in casual conversations or small talk.

To mitigate these limitations, fine-tuning pre-trained models using real conversational data is an effective solution. This can be achieved through supervised fine-tuning (SFT), which fine-tunes the LLMs with dialogue datasets to enhance natural speech patterns or reinforcement learning (RL) \cite{kaelbling1996reinforcement}, which optimizes conversational fluency by incorporating human feedback.
Despite these improvements, users can still distinguish between AI and human-generated dialogue due to several key limitations \cite{fuchs2024understanding,chein2024human}, including 
\begin{itemize}
    \item Lack of genuine understanding and emotional depth.
    \item Inability to express personal opinions or experiences.
    \item Limited personalization and memory consistency, which affects coherence in long-term interactions.
\end{itemize}
To enable the AI systems to exhibit stronger human-like intelligence in conversation, one may need to empower the system to first think like a human.

\subsection{Thinking like a Human}

A typical structure of conversational AI systems is the ``prompt+model call'' style, where comprehensive instruction with persona, language style, and few in-context examples are inserted into the system prompt in addition to the dialog history (with new user input), and model to generate contents as instructed.

However, it is hard for such an approach to exhibit human-like thinking and reasoning in the conversation, as the stronger an LLM’s instruction-following ability, the less dynamic its responses become. 
This occurs because LLMs generate text through sequential token prediction, selecting the most probable next word based on previous inputs. For example, if a prompt instructs an LLM to "show proactivity in conversation," it may overuse questions to create an illusion of engagement. 
Similarly, if a prompt emphasizes "conciseness," the LLM may generate overly brief and rigid responses. 
Therefore, prompting or in-context learning might be insufficient to guide such thinking and generating process.

Therefore, for AI to think and reason in an anthropomorphic way, it must not only predict language but first understand the underlying cognitive processes that drive human conversations, and a single model call of a foundational LLM might be insufficient to accomplish such a task.

\subsection{Road-map and Our Contributions}

To address these cognitive limitations, there are two tasks involved. Firstly, we need to fully understand the key aspects of human intelligence involved in conversations. And secondly, to better human likeness, we need to involve human factors in the loop, such that fine-tuning techniques such as reinforcement learning can be leveraged. 

Therefore, we propose a two-stage solution for developing anthropomorphic conversational AI.
Regarding the first stage, we would like to investigate the recursive thinking-expressing loop of humans and the key aspects involved in the loop. Then, we establish a framework with foundational LLMs that replicates those key aspects, such that enhanced anthropomorphism could be exhibited by a system implemented by the framework.
In the second stage, given the conversation records and extensive thinking and reasoning generated by the framework, with reasonable labeling of human preference, we may be able to fine-tune (e.g., reinforcement learning techniques such as human-feedback reinforced learning (HFRL) \cite{griffith2013policy}, proximal policy optimization (PPO) \cite{schulman2017proximal}, direct preference optimization (DPO) \cite{rafailov2023direct}, group relative policy optimization (GRPO) \cite{shao2024deepseekmath}) foundational LLMs to enhance conversational intelligence capabilities.

In this work, we focus on the first stage of the solution, more specifically, we make the following major contributions,
\begin{itemize}
    \item We carefully analyze the key aspects that may enhance the anthropomorphism in conversational AI systems.
    \item We propose a multi-module framework in which each module manages one (or more than one, sometimes) key aspect(s) of the key modules, and a workflow that comprises the modules to implement a human-like reasoning-expressing loop.
    \item We design a two-step real-person involved experiment, with human-generated test conversations and evaluations, it is verified that our framework could significantly improve the conversation experience even without human preference-involved training on the LLMs.
\end{itemize}

%% file: sections/preliminaries.tex
\section{Towards Anthoropomorphism}\label{sec:preliminaries}

The concept of anthropomorphism, which involves attributing human characteristics to non-human entities, has been a central focus in the development of conversational AI. This approach emphasizes equipping AI systems with fundamental personification, social intelligence, and conversational intelligence, which has been extensively explored in the literature.

Personification focuses on assigning the AI system personalized characteristics such as personality and identity, which is essential for anthropomorphic applications. 
Conversational Intelligence, involves basic mechanisms such as thinking, memorizing, and learning, enabling individuals to acquire and understand information about themselves and others from the conversation, and manage the conversation's trajectory. 
Social Intelligence refers more specifically to the immediate perception of both internal states and the external environment which leads to a deeper understanding of emotion and social cues.

\subsection{Personification}

\vspace{5pt}
\noindent\textbf{Personality and Identity}

This notion primarily refers to a set of configurations that define a virtual (AI) character through various settings. 
For example, an AI assistant might be programmed with a methodical thinking mode, which emphasizes logical, structured reasoning for problem-solving. Its language style could be tailored to be both professional and engaging, using precise yet approachable vocabulary to effectively communicate with users. Further defining its personality, the AI might include a programmed sense of humor and the capacity for emotional intelligence, enabling it to understand and react to human emotions. Additionally, incorporating cultural sensitivity allows the AI to interact in ways that resonate culturally, making the experience more relatable and enriching for users from diverse backgrounds.

\subsection{Conversational Intelligence}

The category \textit{conversational intelligence} includes key elements that enable AI to understand each others' perspectives and engage in meaningful interaction. 
There is a rich collection of studies regarding different aspects of this category, such as natural language understanding, intent recognition, etc.
Here, we view the problem from a more practical perspective and focus on several key aspects that dominate the user's conversational experience.

\vspace{5pt}
\noindent\textbf{Context Understanding}

As the fundamental of successful conversations, the AI needs to first correctly understand the context, which consists of the following key elements:
1) The content of the conversation, such as theme, agenda, topic, and any underlying subject-specific knowledge relevant to the user's query. Context comprehension requires identifying explicit and implicit cues within the conversation to form an accurate understanding of the subject matter.
2) The logical relationship and consistency. This involves tracking the conversational trajectory, maintaining coherence, and ensuring responses align with prior statements. The AI must identify contradictions, logical inconsistencies, or shifts in context to preserve the flow of interaction.

Context understanding enables the AI to effectively manage dialogue and respond appropriately to user inputs, which may include ambiguous phrasing, incomplete sentences, or colloquial expressions. 
This capability is essential for ensuring coherence and aligning the interaction with the user’s intent.

\vspace{5pt}
\noindent\textbf{Reflexive and Analytical Thinking}

To further enhance conversational quality, AI systems may need both reflexive and analytical thinking. 
Reflexive thinking enables the AI to quickly respond to user inputs. 
For straightforward queries, a brief chain of reasoning is often sufficient to generate an appropriate reply. 
Even in cases where user inputs are complex, reflexive thinking allows the AI to provide an initial response, serving as a buffer for deeper analysis.
Analytical thinking, by contrast, equips the AI to handle complex queries by processing intricate details, synthesizing information, and formulating well-reasoned answers. 
Together, these complementary abilities create a conversational experience that is both highly responsive and intellectually robust.

\vspace{5pt}
\noindent\textbf{Linguistic Competence}

As another fundamental element, \textit{linguistic competence} permits the AI to accurately turn thoughts into written or verbal-styled languages and deliver them to the user. 
This involves not only grammatical accuracy and vocabulary richness but also the ability to tailor language to match the personality and identity assigned to the AI (which will be discussed in greater detail later). 
Additionally, linguistic competence enables the AI to detect and adapt to different communication styles, enhancing the user's overall experience by ensuring clarity and engagement.

\vspace{5pt}
\noindent\textbf{Proactivity}

Proactivity refers to the AI's ability to anticipate user needs or the trend of the conversation, identify potential areas of interest or concern, and take initiative in steering the conversation toward users' preferences (e.g. user's satisfaction in free chats or desired meaningful outcomes in a task-oriented conversation). 
This involves suggesting relevant topics, asking clarifying questions, or offering recommendations based on prior interactions and contextual understanding. 

By demonstrating proactivity, the AI can create a more dynamic and engaging experience, ensuring the user feels understood and supported while minimizing the need for repetitive inputs.

\vspace{5pt}
\textbf{Memorization}

Memorization involves the AI’s ability to retain and utilize information from prior interactions, fostering continuity and personalization. 
Effective memorization enables the AI to:
1) Enhance Personalization: Recall user preferences, history, or specific details to tailor responses.
2) Maintain Conversational Flow: Reference previous interactions to avoid redundancy and maintain coherence across sessions.
However, it is worth noting that implementing memorization raises challenges regarding data privacy, storage, and ethical considerations, necessitating robust mechanisms to safeguard user information while delivering a seamless conversational experience.

\subsection{Social Intelligence}

\textit{Social Intelligence} usually refers to the capability of managing social behavior and activities effectively and adaptively in various interpersonal contexts. 
From the perspective of AI, social intelligence is fundamentally grounded in an awareness of both the AI's state and the other party's state, encompassing the ability to understand and manage emotions, interpret social cues, and adapt to the dynamic interplay of interpersonal interactions.

The state of the conversation, such as the attitude (curiosity, resistance, or interests) of both parties regarding the conversation, including subtle emotional nuances and tone inferred from linguistic cues, plays a central role in shaping AI's social intelligence. 
These elements enable AI to better navigate interpersonal dynamics and foster meaningful and cooperative interactions.

\vspace{5pt}
\noindent\textbf{Other/Self-Awareness}

Social intelligence begins with awareness - for both the AI's internal state and its perception of the other party in the conversation. 
Here, self-awareness involves the recognition of AI's own role, feelings, and objective in the conversation. 
This includes an understanding of how its responses might influence the conversation and the broader goals of the interaction.
On the other hand, other-awareness involves the ability to perceive and interpret the user’s perspective, emotions, attitudes, and intentions.  
This information can be gleaned by analyzing linguistic features such as word choice, tone, syntax, and non-verbal cues when available, such as pauses or response timing.

By synthesizing these two forms of awareness, the AI can better align its actions with the interaction’s objectives, ensuring responses are not only contextually appropriate but also sensitive to the user's emotional and social state.

%% file: sections/system.tex
\section{Framework}\label{sec:system}

\begin{figure}
    \centering
    \includegraphics[width=0.8\linewidth]{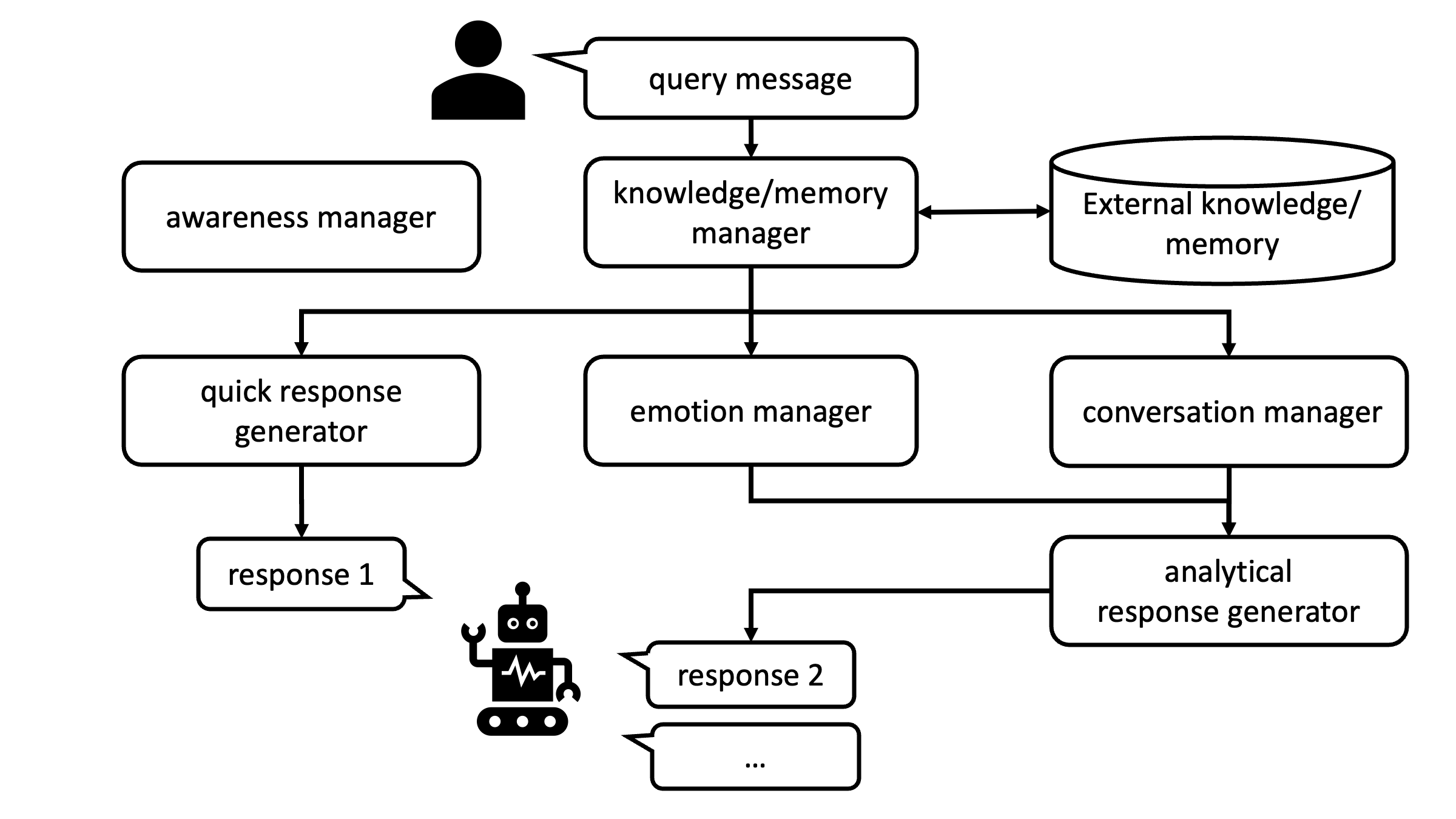}
    \caption{The workflow of the framework. Here, the awareness manager is called after agent complete all the outputs and focus on the self-awareness (opinion, feeling and emotion), while emotion manager focuses on the control of emotion in respond to the user input. The conversation manager focuses on the strategical analysis of the current topic or task of the conversation. The quick response generator is designed to provide quick response for simple user input, or provide buffer time for the agent to process complex user input. Note that the analytical response generator is involved in a loop - it would analyze the existing response messages, and decide whether to continue to output messages, or conclude the current turn.}
    \label{fig:system}
\end{figure}

In Section~\ref{sec:preliminaries}, we have introduced the key characteristics that need to be addressed in an anthropomorphic conversational AI system. In this section, we propose a practical framework that handles the characteristics.

\subsection{Overview}

\vspace{5pt}
\noindent\textbf{LLM-based Multi-module Framework}

Before diving into the details, we pose an essential question: \textit{Is a single LLM call sufficient to generate anthropomorphic responses?} 
This question arises from the limitations observed in many existing conversational AI solutions, which usually consist of a prompt with detailed instructions (including the personality and language style settings, some examples), and a LLM (which could be tuned for finer generation results).

While the solutions could respond in human (spoken language) tones, they often fail to show intelligence in both conversational and social perspectives. 
For example, regarding the proactivity (dynamism) in the conversations, they often show either a lack of proactivity (the user needs to initiate and sustain dialogue) or mechanical proactivity (e.g., continually asking questions). 

There could be a dilemma when one wants to generate the response with dynamism within one model call. 
On one hand, one can prompt the model to show some dynamism, but the models may fail to understand the meaning of "dynamism" in the context.
On the other hand, one can add more instructions regarding the dynamism in detail, but it may consist of many rules, resulting in either the model failing to follow the rules (as there might be many of those), or making results look mechanical and lack genuine engagement. 
Over time, this can lead to user disengagement, as continually formulating or answering questions can become burdensome. 

Our findings suggest that \textit{when the LLMs are not sufficiently powerful, multi-calls are often a more effective approach for achieving anthropomorphism in conversational AI}. Below, we outline the key reasons for this proposition.

\begin{itemize}
    \item \textit{Better Response Time}. As discussed in the previous section, enabling an AI system to exhibit anthropomorphic characteristics requires comprehensive information extraction from both dialogue history and memory. Executing multiple complex tasks in a single model call increases the number of input and output tokens, leading to longer processing and response times. By executing tasks in parallel, we can reduce response time, which is crucial for improving the overall user experience.
    \item  \textit{Independent Thoughts}. Due to the underlying mechanism of LLMs, the prediction of each token depends on all input and previously generated tokens. For tasks such as self and other-awareness extraction, it is necessary to perform these operations independently. However, unless the models have exceptional instruction-following capabilities (which usually means they are large enough in scale), we find that the results of one task often interfere with those of another. By dividing tasks into separate model calls or functional modules, this interference can be minimized, which is also proven in our tests.
\end{itemize}

\vspace{5pt}
\noindent\textbf{Implementation of the Intelligence Characteristics}

In Section~\ref{sec:preliminaries}, we have introduced several key characteristics of anthropomorphism in conversational AI. Now, we implement the characteristics in our framework which consists of multiple specialized modules (model calls wrapped with needed instructions, information, and other resources), each handling distinct intelligent characteristics in an efficient and effective way.

To wrap and implement the relevant characteristics into modules, a potential solution is to imitate human thoughts in conversation, which may pose another question: \textit{what are the underlying thoughts that involve the characteristics, influence, and generate human responses?}

Categorized by the stage in the chain of thoughts (in a conversation), the thoughts should include: 1) basic information related to the personality and identity aspects, 2) the analysis of the conversation regarding the agent and the user, and 3) the logic of generating responses with the other thoughts.
We need to address that 2) and 3) involve both the conversational and social intelligence characteristics as introduced in Section~\ref{sec:preliminaries}, which are heavily twisted in the thinking steps.

\subsection{Memory Manager}

Memory is perhaps the most common aspect of conversational intelligence and has an impact on nearly every part of the framework. 
In our approach, memory management consists of two parts: internal memory and external memory. Internal memory includes information directly related to the AI system, such as dialog history and memory pieces, while external memory refers to knowledge resources such as databases and online sources.

\vspace{5pt}
\noindent\textbf{Dialog History} module collects the user's query and the agent's responses and provides the required length of history as requested by all other modules.

\vspace{5pt}
\noindent\textbf{Internal Memory Manager} assists the agent to manage the stored memory corresponding to the agent itself and the user. The stored memory consists of the memory pieces extracted from the conversation (user's memory) and the memory regarding the AI character (agent's memory). It performs two main tasks:
\begin{itemize}
    \item Retrieving the most relevant information from the internal memory based on the user's input.
    \item Extracting useful memory pieces (e.g., events, personal relationships, user preferences) from the dialog history and storing them in memory. The manager periodically merges duplicates, resolves conflicts, and reorganizes memory pieces during the conversation.
\end{itemize}

\noindent\textbf{External Knowledge Manager} manages the online (e.g. web search) and offline (e.g. database) knowledge sources. It may utilize retrieval augmentation generation (RAG) techniques such as query rewrite, multi-source retrieval, and summarization to provide the agent with the most relevant external knowledge.

\vspace{5pt}
One may notice that the stored memory consists of the user's memory and the agent's memory. While many vanilla applications may inject the memory (dialog examples, or the story of the character) into the prompt, in practice we found that it has significant shortages. First, a detailed story or comprehensive examples means a large number of input tokens. Second and more importantly, it may cause stereotypical behavior like repeatedly using phrases or expressions in the examples or mentioning certain facts in the story.

While the obligation of Dialog History and Internal Memory seem to overlap with each other, they play very different roles in the system. On one hand, the Dialog History is usually taken as part of the prompts, therefore it is necessary to limit the length of history to be taken. Although the current LLMs can take long contexts as input, the extraction of information is also burdensome for the LLM, since it is not the only task that needs to be handled. On the other hand, the Internal Memory is designed to save and retrieve the extracted information, which significantly reduces the workload for each memory retrieval. Moreover, the extraction process also labels the information pieces (by reforming them into designated format instead of original messages) to make the memory storage more organized. While the conversation proceeds, the preference information from the user accumulates, which will allow the agent to provide a more personalized chat service.

\subsection{Awareness Managers}

As previously discussed, awareness plays a key role in the system. We propose separate components to manage self- and other-awareness, while each of them would manage both the conversational and social intelligence characteristics. 

\vspace{5pt}
\noindent\textbf{Self-Awareness Manager} is responsible for managing the AI agent’s awareness, which is vital for both social and conversational intelligence. Guided by the Personality and Identity, this manager analyzes the Dialog History to generate or extract the following information about the agent:
\begin{itemize}
    \item Conversational perspective: 1) the agent’s satisfaction and opinions on the current topic, 2) the topic the agent finds most interesting, and 3) the agent’s planned course of action (e.g., explore further, switch topics, or stop). 
    \item  Social perspective: 1) the agent’s current emotion, 2) the emotion the agent is likely to express next, and 3) the tone and language style that best reflects the emotion. 
\end{itemize}

\noindent\textbf{Other-Awareness Manager} functions similarly to the Self-Awareness Manager but focuses on the user’s perspective. Also guided by the Personality and Identity, it analyzes the Dialog History to extract or generate the following information about the user: 
\begin{itemize}
    \item Conversational perspective: 1) the meta-topic of the conversation, 2) the user’s satisfaction with the current topic, 3) topics that may interest the user beyond the current discussion, 4) identify whether the conversation is task-oriented, if so, what is the best step-by-step strategy to accomplish the task, what is the current stage in the strategy, and what should be done next.
    \item Social perspective: 1) the user’s current emotion, and 2) the most natural emotional response to the user’s current state.
\end{itemize}

One may notice that the obligations of the awareness managers explicitly mention the judgment on whether the conversation is task-oriented. This is crucial for the analytic generator (which will be introduced later), as the strategy mentioned in the obligations plans a chain of actions (or thoughts) for the generator, and explicitly instructs the generator about the current position on the plan. This would better generate desired result, and push the conversation towards the goal (task-completing).  

Additionally, it is worth mentioning that awareness managers play as the prime providers of conversational proactivity.
When the topic interests either the agent or the user, the manager may suggest further exploring the topic by showing proactivity such as asking questions or expressing interest. 
If the topic is not interesting for either party, based on the personality of the agent, it may suggest to either stick to the topic or switch to a different one. 
If the atmosphere of the conversation is hostile, the agent may chose to conclude the conversation.
These and all other uncertainties together form a dynamic instruction (prompt) for the response generators (which will be discussed shortly), thus implement the dynamism in the conversation.

While it is possible to implement the awareness managers in a single model call, we recommend splitting their tasks into separate calls and executing them in parallel. This approach offers two key benefits: 1) Simpler tasks allow for the use of smaller models, which reduces costs and leads to shorter response times (especially when executed in parallel).
2) Despite the similarities between the tasks of the two managers, conflicts can arise due to differing perspectives (agent vs. user). Executing the tasks separately helps prevent such conflicts and can enhance the dynamism of responses, reflecting the complexity of human thought.

Therefore, we suggest executing these tasks independently. Additionally, we have found it beneficial to incorporate the Self-Awareness Manager as a "re-think" phase at the end of the workflow, after all responses have been generated, to better mirror the psychodynamical inertia of the agent.

\subsection{Response Generators}

Inspired by natural human reactions in interpersonal conversations, our framework incorporates two response modes: reflexive and analytic. The reflexive mode handles simple queries that do not require deep thinking, while the analytic mode processes more complex queries, using all relevant information for comprehensive responses. 

\vspace{5pt}
\noindent\textbf{Quick Response Generator} considers the Personality and Identity, analyzes the Dialog History, and uses Self-Awareness information to generate an initial response to the user’s input.

\vspace{5pt}
\noindent\textbf{Analytical Response Generator} is activated after the reflexive response and builds on the previous message. This generator incorporates External Knowledge and Other-Awareness information (which may include instructions) to create a more in-depth, thoughtful response.

In terms of the conversational intelligence, the two generators are designed to implement reflexive and analytical thinking characteristics, respectively, while the linguistic competence are mainly governed by the capability of LLM (which can be fine-tuned of course).

While the two generators have clearly defined roles, we would like to share some insights regarding the workflow design. 
As shown in Fig.~\ref{fig:system}, the analytic generator should be placed in a loop with exit conditions. These conditions can be divided into two components: one purely stochastic, such as a random selection, and the other a rethinking process that determines whether the generated responses (including both reflexive and analytic responses) adequately address all requirements.
As mentioned earlier, the requirements include suggestions and analysis based on the dialog history, user input, task completion strategy, and other relevant factors. This rethinking process is crucial for an anthropomorphic conversational agent, as the "think-response-rethink-response" cycle mirrors how humans respond to complex queries.

%% file: sections/experiments.tex
\section{Experiment and Evaluation}\label{sec:experiment}

\subsection{Experiment Design}

We use the proposed framework as introduced in Section~\ref{sec:preliminaries} to build a conversational AI service (referred to as the ``agent'' in the following context) with the Qwen-max model as the underlying LLM. We need to address that Qwen-max is a general close-sourced foundation model, and we did not apply any fine-tuning technique to improve the capability of the model in generating spoken-styled languages. The main purpose is to eliminate the impact of fine-tuning in generating anthropomorphic responses. 

To evaluate the performance of our proposed framework, we conducted a two-phase experiment.

\vspace{5pt}
\noindent\textbf{Phase 1}. 
We recruited $6$ volunteers from our firm—$3$ males and $3$ females, aged between $26$ and $35$, representing various occupations—to engage in $20$-minute conversations with the AI agent. We collected all dialog histories, segmenting them into samples with a sliding window of width $20$ messages (combining user input and agent response messages). This process resulted in a total of $340$ conversation samples. We randomly selected $5$ samples from this pool to form a set, independently repeating this process $30$ times to create $30$ test sets.

\vspace{5pt}
\noindent\textbf{Phase 2}. 
Other than the conversation participants, we recruited another group of $12$ volunteers from our firm, ensuring an even distribution of genders and occupations, to serve as evaluators. We designed a questionnaire featuring $8$ statements and one open question, evenly split between assessments of conversational intelligence and social intelligence. All evaluators were required to 1) independently and randomly pick $3$ sample sets and carefully read the samples, 2) based on their agreement with each statement, give a rate on a scale from $1$ (strongly disagree) to $7$ (strongly agree), 3) answer the open question.

\textbf{Remarks:} We need to address that we haven't involved large-scale human evaluation and ablation study regarding various LLMs in this report. This report mainly focuses on presenting the framework design, which has been deployed in real-world applications. In a later version, we would like to conduct more comprehensive experiments, please stay posted.

\subsection{Assessment Criteria}

The statements of the questionnaire include:
\begin{enumerate}
    \item The flow of conversation was smooth and consistent; for example, the AI followed well with the flow of the conversation, did not misunderstand references to past contexts, and the thoughts expressed were natural and coherent.
    \item The AI demonstrated intelligent proactivity in conversations, such as expressing feelings, opinions, perspectives, or curiosities, rather than mechanically responding or querying the user.
    \item The AI effectively managed topic transitions initiated by the user, showing natural reactions and adapting seamlessly to changes.
    \item The AI exhibited linguistic competence, using appropriate language across various scenarios to respond, express emotions, or convey opinions.    
    \item The AI communicated its feelings and emotions effectively; for instance, emotional shifts in the AI’s responses were noticeable.
    \item The AI expressed its interests and preferences in conversations, such as by proactively asking questions or employing other methods to demonstrate engagement.
    \item The AI accurately captured and reflected the user's emotions in conversations, recognizing emotional changes in the user and responding appropriately.
    \item The AI acknowledged and explored the user's interests and preferences, delving deeper into topics the user showed enthusiasm for.
\end{enumerate}

Here, Statements $1$ to $4$ relate to conversational intelligence, while the other four focus on social intelligence.
In the end, we asked an open question, \textit{What do you think are the pros and cons of the AI's performance in the conversation samples?}

\subsection{Result Analysis}

One may notice that some characteristics in Section~\ref{sec:preliminaries} are not explicitly mentioned in the statements. This is mainly because those factors implicitly affect many observations (for example, the context-understanding ability is fundamental for almost all the statements), or are hard to evaluate directly (e.g., memorization is essential for the agent to correctly understand context when referring to the past conversation, but the samples may not always contain dialog referring to the past).

\begin{figure}[ht]
    \centering

    \begin{subfigure}[t]{0.4\textwidth}
    \centering
        \includegraphics[width=\textwidth]{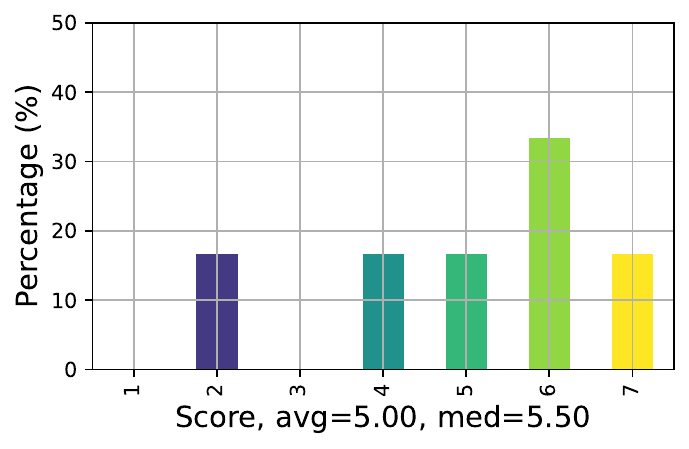}
        \caption{Conversation management: continuity and fluency.}\label{fig:subfig6}
    \end{subfigure}
    \hspace{1em}
    \begin{subfigure}[t]{0.4\textwidth}
    \centering
        \includegraphics[width=\textwidth]{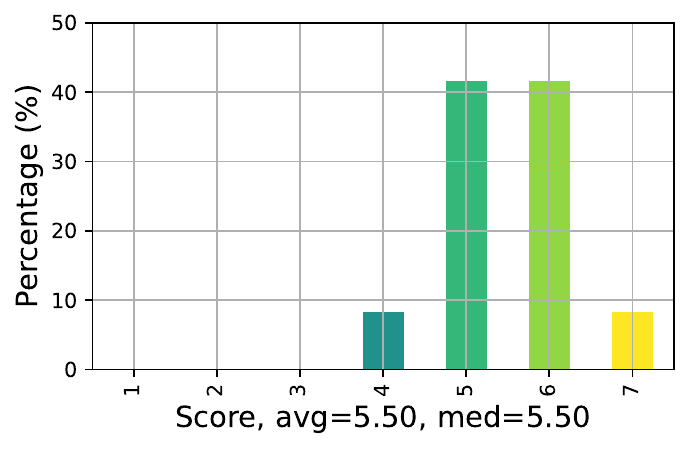}
        \caption{Conversational management: proactivity.}\label{fig:subfig7}
    \end{subfigure}

    \vspace{1em}
    
    \begin{subfigure}[t]{0.4\textwidth}
    \centering
        \includegraphics[width=\textwidth]{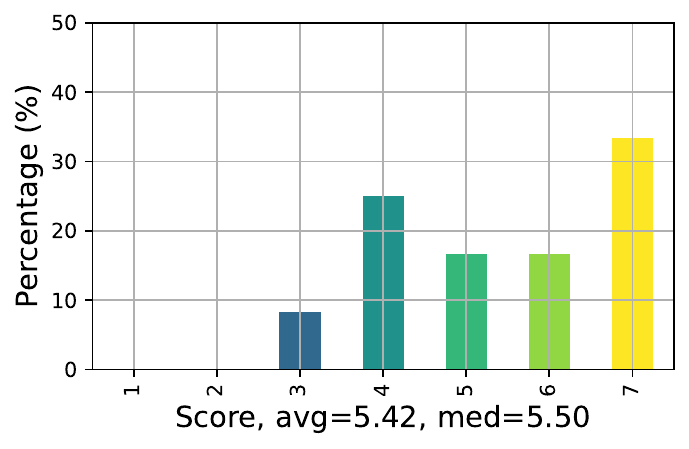}
        \caption{Conversation management: topic transition.}\label{fig:subfig13}
    \end{subfigure}
    \hspace{1em}
    \begin{subfigure}[t]{0.4\textwidth}
    \centering
        \includegraphics[width=\textwidth]{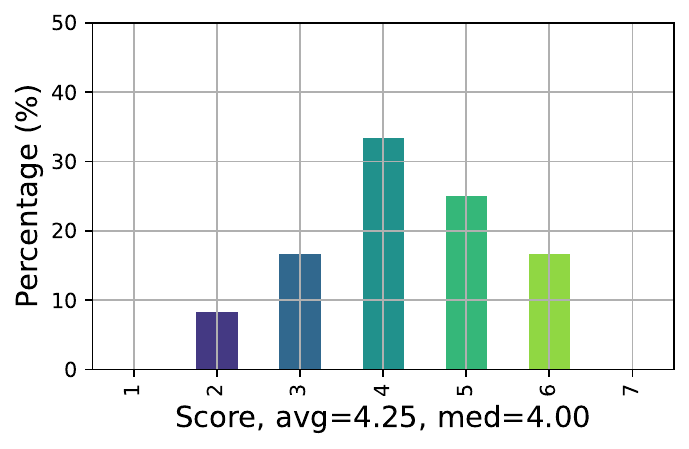}
        \caption{Linguistic competence.}\label{fig:subfig11}
    \end{subfigure}
    
    \caption{Conversational Intelligence}\label{fig:exp_conversational}
    
\end{figure}

\begin{figure}[ht]
    \centering

    \begin{subfigure}[t]{0.4\textwidth}
    \centering
        \includegraphics[width=\textwidth]{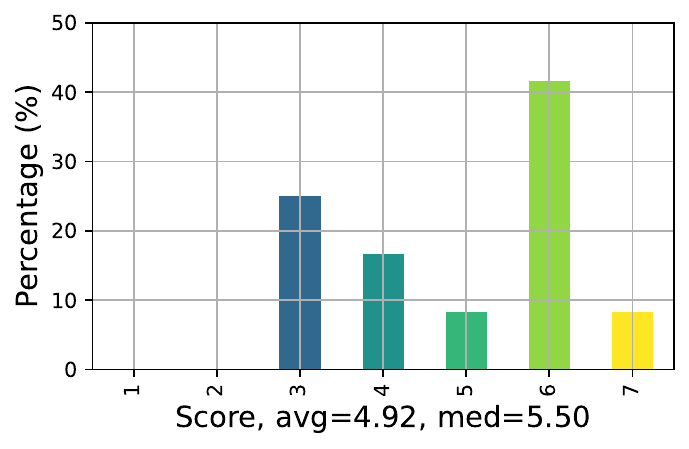}
        \caption{Self-emotion expression.}\label{fig:subfig8}
    \end{subfigure}
    \hspace{1em}
    \begin{subfigure}[t]{0.4\textwidth}
    \centering
        \includegraphics[width=\textwidth]{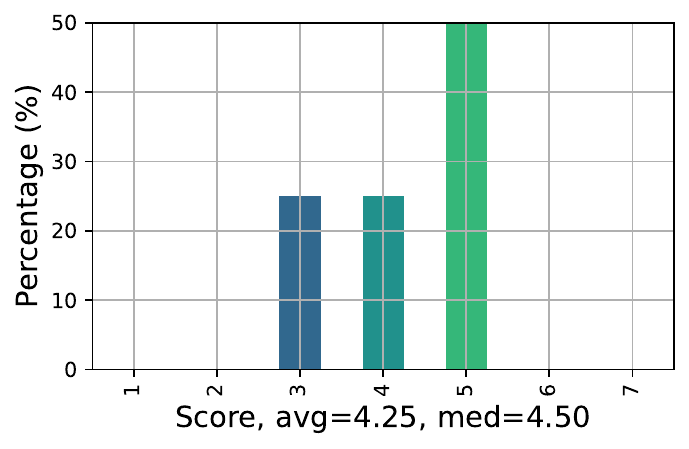}
        \caption{Self-preference (thoughts) expression.}\label{fig:subfig14}
    \end{subfigure}
    
    \vspace{1em}

    \begin{subfigure}[t]{0.4\textwidth}
    \centering
        \includegraphics[width=\textwidth]{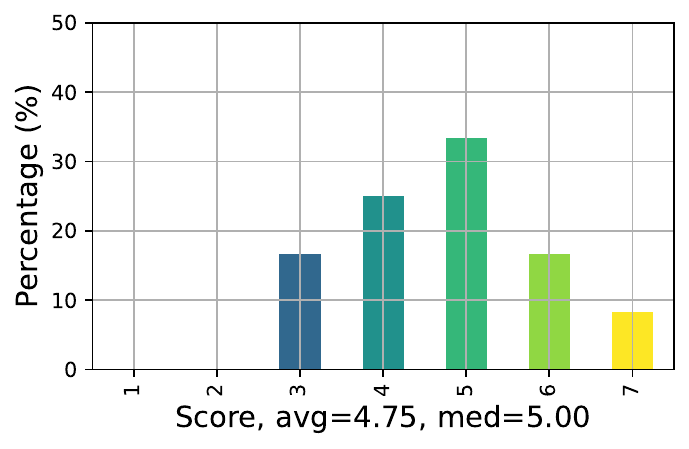}
        \caption{User-emotion reflection.}\label{fig:subfig9}
    \end{subfigure}
    \hspace{1em}
    \begin{subfigure}[t]{0.4\textwidth}
    \centering
        \includegraphics[width=\textwidth]{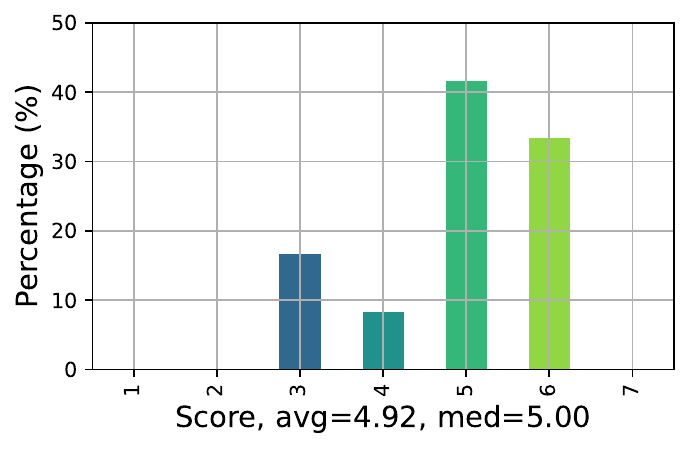}
        \caption{User-preference awareness}\label{fig:subfig12}
    \end{subfigure}
    
    \caption{Social Intelligence}\label{fig:exp_social}
\end{figure}

The rating results are presented in Figures~\ref{fig:exp_conversational} and \ref{fig:exp_social} regarding the conversational and social intelligence factors, respectively.

As we can see in Figures \ref{fig:exp_conversational}(a) and (c), in general, the evaluators give positive ratings regarding the continuity and fluency of the conversation flow, and the agent well responded to the transition of the topics. As discussed earlier, this implicitly implies that context-understanding and memorization abilities are sound enough to handle the conversation flow subject to either continue in-depth or change.

Particularly, the agent displayed a strong ability of dynamic conversational proactivity in Figure~\ref{fig:exp_conversational}(b), which is also mentioned in the feedback from the evaluators that the agent showed interest in particular topics and proactively interacted with the human to move the conversation forward in a dynamic and intelligent way. 

Regarding linguistic competence, the ratings reflected a mixed experience in Figure~\ref{fig:exp_conversational}(d). Not surprisingly, the majority of the critical feedback targeted the language used in the conversation.
While linguistic competence is complex and the rating is highly subjective, the evaluators pointed out in the feedback that the AI often used repetitive phrases and emojis excessively, and used overly structured language in the responses, which made the conversations feel unnatural. 
The formal tone of the AI's language was also a concern, as it did not always suit the casual nature of some dialogues. 
As we mentioned earlier, the LLM we used in the agent is a foundation model without fine-tuning regarding this task, which will be tested in our future work.

Despite these issues, there were some positive aspects. 
In terms of social intelligence characteristics, the AI demonstrated overall strong other- and self-awareness capability in Figure~\ref{fig:exp_social}. The agent successfully captured the user's emotion and expressed its own emotion in the conversation which is reflected in the response and the underlying strategies. The feedback mentioned a case that after receiving insults, the agent appropriately disengaged users, which is a consequence of emotion management engaged in the response generation process.

Also, the overall rating suggests that the agent was able to capture the user's preference reflected in the conversation and express its preference (which is partly configured in the personification settings), in which case the agent may show interests in the preferred topic, or switch the topic when the topic bored either the user or the agent.

%% file: sections/conclusion.tex
\section{Conclusion and Future Works}

In this work, we propose a practical framework for building anthropomorphic conversational AI systems. By utilizing various components to manage the key characteristics in both conversational and social intelligence, users can obtain a human-like conversational experience without effort in fine-tuning the foundational large language models.

In the future, we will move on to the second stage of our anthropomorphic solution. Given the extensive data generated in the framework during test conversations, we would first involve human labeling to capture the human feedback regarding the conversation and corresponding thinking steps, then move on to reinforcement learning to build a language model with much stronger anthropomorphic conversation capability.